\documentclass[runningheads,a4paper]{llncs}
\usepackage{amssymb}
\usepackage{graphicx}
\usepackage{multirow}
\usepackage{amsmath}
\usepackage{wrapfig}
\usepackage{color}
\usepackage{url}
\usepackage{booktabs,tabularx}
\usepackage{todonotes}

\urldef{\mailsa}\path|e.talavera.martinez@rug.nl,|
\urldef{\mailsb}\path|petia.ivanova@ub.edu,|
\urldef{\mailsc}\path|n.petkov@rug.nl| 
\newcommand{\keywords}[1]{\par\addvspace\baselineskip
\noindent\keywordname\enspace\ignorespaces#1}

\begin{document}

\mainmatter  

\title{Towards Egocentric Sentiment Analysis}


%
%
\author{Estefania Talavera%
\and Petia Radeva\and Nicolai Petkov}
\authorrunning{Towards Emotion Retrieval in Egocentric PhotoStream}

\institute{University of Barcelona, Spain, and University of Groningen, The Netherlands\\
\mailsa
\mailsb
\mailsc}

\maketitle

\begin{abstract}
The availability and use of egocentric data are rapidly increasing due to the growing use of wearable cameras.
Our aim is to study the effect (positive, neutral or negative) of egocentric images or events on an observer. Given egocentric photostreams capturing the wearer's days, we propose a method that aims to assign sentiment to events extracted from egocentric photostreams. Such moments can be candidates to retrieve according to their possibility of representing a positive experience for the camera's wearer. 
The proposed approach obtained a classification accuracy of 75\% on the test set, with deviation of 8\%. 
Our model makes a step forward opening the door to sentiment recognition in egocentric photostreams.
\end{abstract}

\section{Introduction}

Lifelogging describes an egocentric vision of the experiences of a person. Nowadays, the use of small wearable cameras, which capture images in certain intervals, is increasing considerably. Such images provide an overview of the daily activities of a person that can be interpreted as a visual log of the day. This information can be used to examine a person's pattern of behaviour; daily habits, such as eating habits, social interactions, indoor or outdoor activities, are recorded in such images. Although our mood is influenced by the environment and social context that surrounds us, egocentric data do not always catch our attention or induce the same emotion when retrieved. We consider that the creation of a diary of positive moments in an electronic way, by combining several cues, will help to improve the inner perception of the user's own life. Therefore, in this work we seek for \textit{positive moments} that can raise the user's positiveness. 

Several experiments have been conducted in that direction, in \cite{Santos2013TheReview.} the authors presented a survey of positive psychology strategies that demonstrated to be potentially effective as tools for the treatment of depression. As an example, in \cite{Beck1986StimulatingConnotation.} the authors suggested to the participants activities such as \textit{walking in a park}, \textit{visiting a friend}, or \textit{going to the Student Union and saying hello to someone}, that resulted in participants' suffering to decline or disappear. Their study gives ideas about the type of moments that can retrieve positive feelings; nature, landscapes, friends, or smiling people staring at us, among others.

Sentiment analysis from images is a novel research field. Given the challenge of image sentiment recognition and the ambiguity of the problem, we analyse images sentiment assigning a discrete ternary sentiment value (positive (1), neutral (0) or negative (-1) value), similar to \cite{You2015RobustNetworks}. In the literature, sentiment recognition from images has been approached based on different types of data features. Attributes such as facial expressions are used for sentiment prediction in \cite{Levi2015EmotionPatterns,Yuan2013SentributeDescriptors}. The combination of visual and textual information from the images \cite{Wang2014MicroblogModel,You2016Cross-modalityMultimedia} appeared due to the wide use of online social media and microblogs. In such websites, images are posted with short descriptive comments by the user. Audio features were also included in the models presented by \cite{Nojavanasghar2016EmoReact:Children,Poria2014FusingContent}.

\begin{figure}[ht!]
\centering
    \includegraphics[width=1\textwidth,height=7cm]{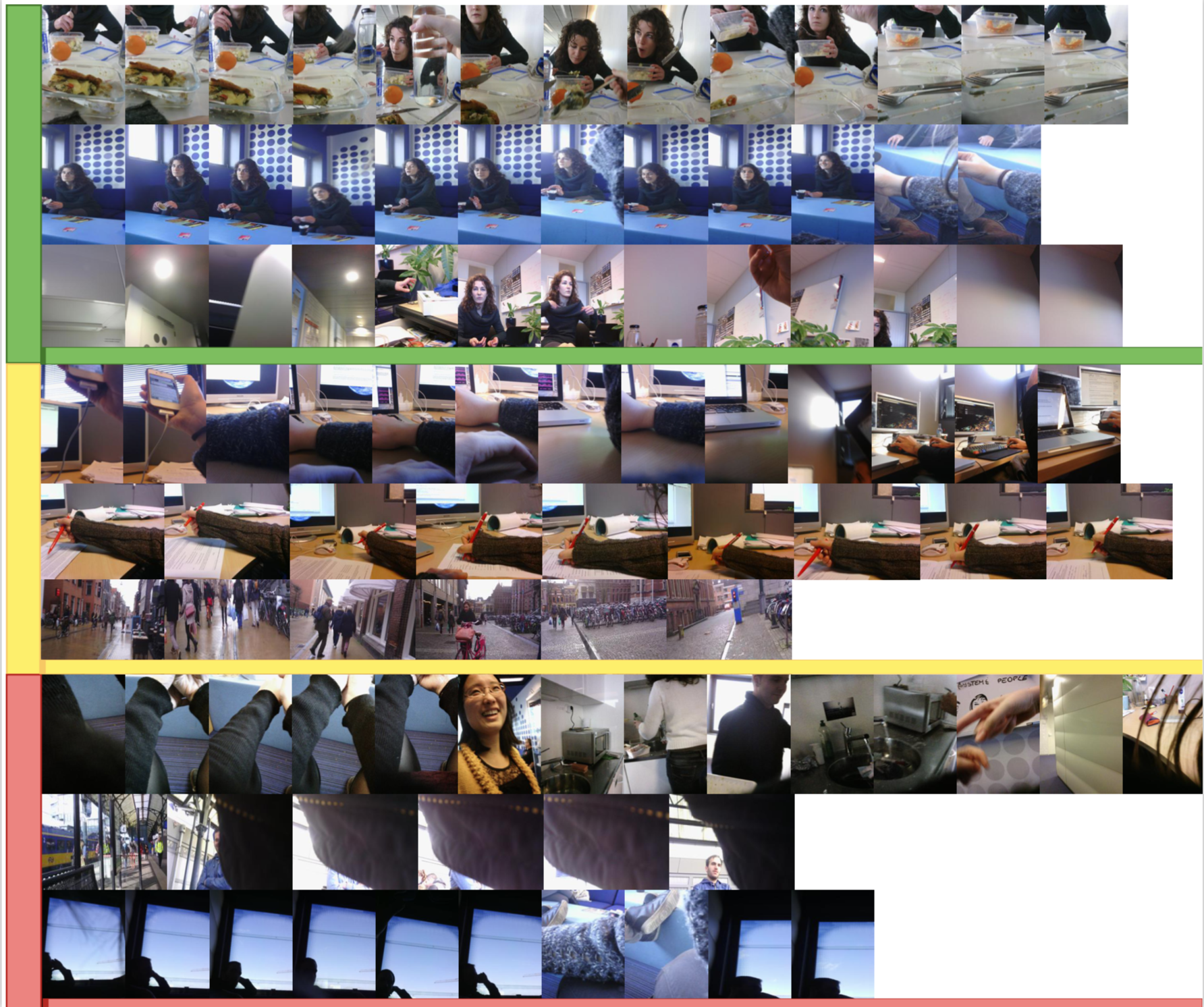}
    \caption{Example of events extracted from the photostream by applying the temporal segmentation method introduced in \cite{Dimiccoli2015SR-Clustering:Segmentation}. Each row of the figure represents an event of the day. Image events labelled as Positive (green), Neutral (yellow) and Negative (red).}
\label{fig:imagesexamples}

\end{figure}

Despite several works having approached the understanding of how people can be affected seeing images, the field of sentiment analysis from images has not been yet settled. The labelling of the images is not yet established, leading to different questions when addressing sentiment recognition from images. As examples, we can find available datasets labelled as: amusement, anger, awe, disgust, excitement, fear, sad \cite{Machajdik2010AffecitveTheory,You2016BuildingBenchmark}, Positive/Negative \cite{Borth2013Large-scalePairs}-Twitter, Positive/Negative/Neutral\cite{Dan-Glauser2011TheSignificance}, with values of Pleasure, Arousal and Dominance \cite{Lang1997InternationalRatings}, with sentiment values from -2 to 2\cite{Borth2013Large-scalePairs}, where the extremes correspond to Negative and Positive sentiments, respectively. Despite the above mentioned works, to the best of our knowledge, none of them has dealt with the sentiment recognition from egocentric photostreams. 

Recently, with the outstanding performance of the Convolutional Neural Networks (CNN), several approaches on sentiment analysis have relied on supervised learning through deep learning techniques, such as \cite{Campos2015DivingPrediction,Levi2015EmotionPatterns,Ma2016GoingRecognition,You2016BuildingBenchmark}. One of the more remarkable approaches, in \cite{Borth2013Large-scalePairs}, introduced a Visual Sentiment Ontology (VSO), based on the Plutchik's wheel of emotions \cite{Plutchik.1980Emotion:Synthesis}, and a visual concept detector called SentiBank. The VSO is built by 3022 semantic concepts called Adjective Noun Pairs (ANP) represented by images from the social net Flickr with them as tags. The ANPs are composed by a pair of a noun and an adjective, with a sentiment value associated between [-2 : 2]. They defend that an object, according to its appearance has a different sentiment value associated to it, like 'lonely boat' (-1.43) and 'traditional boat' (1,37), or 'noisy bird' (-1) and 'cute bird' (1,37). They proposed the semantic concepts baseline classification based on visual features (RGB, SIFT, LBP, etc) extracted from the images. In \cite{Chen2014DeepSentiBank:Networks}, a Deep Neural Network named DeepSentiBank was trained on Caffe for the VSO semantic concepts classification. The authors relied on the concepts with higher number of images and with a classification accuracy associated. From the original 3022 concepts in \cite{Borth2013Large-scalePairs} they select 2089 in \cite{Chen2014DeepSentiBank:Networks}. 

To the best of our knowledge, previous to our work \cite{Talavera2017SentimentPhotostreams} there was no approaches addressing sentiment recognition from egocentric photostreams. We propose to analyse the output of the DeepSentiBank per image as semantic representation. We defined a classification model where the one-vs-all SVM classifiers were trained and evaluated with the features describing semantic and global information from the images. 

In this work, we approach the same problem from a different perspective. We analyse the relation to each other semantic concepts extracted from images that belong to the same scene. A scene is described as a group of sequential images related between them and describing the same event. Our contribution is an analytic tool for positive emotion retrieval seeking for events that best represent a positive moment to be retrieved within the whole set of a day photostream. We focus on the event's sentiment description where we are observers without inner information about the event, i.e. from an objective point of view of the moment under analysis. 

The rest of the paper is organized as follows. In Sect. 2, we describe the sentiment analysis method and the features selection procedure. In Sect. 3, we describe the proposed dataset, while in Sect. 4, we describe the experimental setup, the evaluation, and discuss our findings. Finally, Sect. 5 draws conclusions and outlines future lines of work.

\section{Method}

\begin{figure}[ht!]
\centering
    \includegraphics[width=1\textwidth,height=6cm]{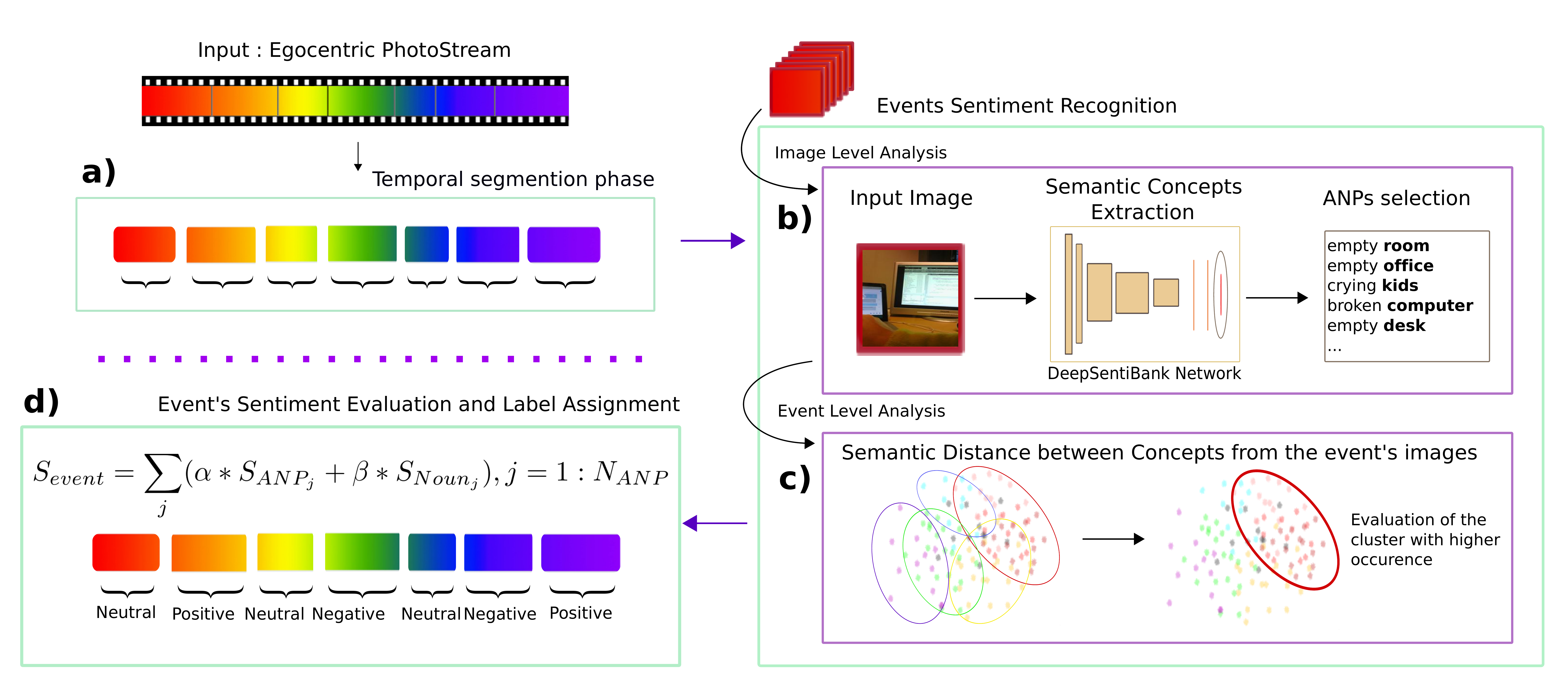}
    \caption{Sketch of the proposed method. First, a temporal segmentation is applied over the egocentric photostream \textbf{(a)}. Later, semantic concepts are extracted from the images using the DeepSentiBank \cite{Chen2014DeepSentiBank:Networks} \textbf{(b)}. The semantic concepts with higher occurrence are selected as event descriptors \textbf{(c)}. Finally, the ternary output is obtained by merging the sentiment values associated to the event's semantic concepts \textbf{(d)}.}
\label{fig:SketchModel}
\end{figure}

Given an egocentric photostream, we propose scene emotion analysis seeking for events that represent and can retrieve a positive feeling from the user. We apply  event-based analysis since single egocentric images cannot capture the whole essence of the situation. By combining information from several images that represent the same scene, we get closer to a better understanding of the event.

\subsection{Temporal segmentation:}
\label{Temporalsegmentation}

We apply temporal segmentation on the egocentric photostreams using the proposed method in \cite{Dimiccoli2015SR-Clustering:Segmentation}. The clustering procedure is performed on an image representation that combines visual features extracted by a CNN with semantic features in terms of visual concepts extracted by Imagga's auto-tagging technology\footnote{http://www.imagga.com/solutions/auto-tagging.html}. In Fig. \ref{fig:imagesexamples} we present some examples of events extracted from the dataset, we introduce below.

\subsection{Event's sentiment recognition:}
\label{Semanticrecognition}

The model relies on semantic concepts extracted from the images to infer the event sentiment associated. However, it relies not only on the semantic concepts extracted by the net with their sentiment associated, but also on how those semantic concepts can be interpreted by the user. We apply the DeepSentiBank Convolutional Neural Network\cite{Chen2014DeepSentiBank:Networks} to extract the images semantic information since it is the only introduced model that extract semantic concepts (ANPs) with sentiment values associated. Given an image, the output of the network is a 2089-D feature vector, where the values correspond to the ANPs likelihood in the image. 

Besides taking into account the sentiment associated to the ANPs, the influence of the common concepts within an event are also analysed. We categorize the noun into Positive, Neutral or Negative. There is a wide range of semantic concepts within the ontology, but many of them seem to repeat concepts that even from the user perspective would be difficult to differentiate when looking at an image; such as "girl" from "woman" or "lady".  

When facing our egocentric images challenge, the VSO presents several drawbacks. On one hand, this tool is trained to recognise up to 2089 concepts, which can not describe all possible scenarios. On the other hand, despite including that big amount of concepts, many of them categorize objects into categories difficult to visually interpret or differ by the human eye. Examples can be the distinction between 'child', 'children', 'boy', or 'kid' from an image. In order to overcome this problem, we generate a parallel ontology with what we consider an egocentric view of the concepts, i.e., we cluster the concepts a person would merge based on their semantic.

\textit{Egocentric analysis of the VSO:} We cluster the semantic concepts based on the similarities between the noun components of the ANPs, which are computed using the wordNet tool\footnote{ http://wordnet.princeton.edu}. Following what would be considered as similar from an egocentric point of view, we manually refine the resulted clusters into 44 categories. We label the clusters as Positive, Neutral or Negative. In Table \ref{table:egoview_vso} we present some of the egosemantic clusters.

\begin{table}[ht!]
\centering
\caption{Examples of clustered concepts based on their semantic similarity, initially grouped following the distance computed by the WordNet tool.}
\label{table:egoview_vso}
\resizebox{\columnwidth}{!}{%
\begin{tabular}{ccc|ccc|ccc}
\multicolumn{3}{c|}{Positive} & \multicolumn{3}{c|}{Neutral} & \multicolumn{3}{c}{Negative} \\ \hline
petals & christmas & award & car & study & bible & tumb & bug & nightmare \\
rose & winter & present & cars & science & book & tumbstone & bugs & accident \\
flora & snow & honor & machine & history & card & monument & insect & shadows \\
park & santa & gift & vehicle & economy & stiletto & grave & worm & noise \\ \cline{2-2}
yard & sketch & heroes & rally & market & sins & memorial & cockroach & scream \\ \cline{3-3} \cline{8-8}
plant & cartoon & dolls & train & industry & record & stone & decay & night \\ \cline{5-5}
garden & drawing & dolls & competition & statue & paper & graveyard & garbage & darkness \\
 & comics & toy & race & sculpture & poem & cementery & trash & shadow \\
 & illustration & toys & control & museum & interview & grief & shit &  \\
 & humor & lego & metal &  &  & pain &  & 
\end{tabular}
}
\end{table}

\subsection{Sentiment Model:}
\label{sentimentmodel}

Given an event, the event’s sentiment analysis model (see Fig. \ref{fig:SketchModel}) performs as follows; 
\begin{enumerate}

\item Given the ego-photostream we apply the temporal segmentation, analyse events with a minimum of 6 images, i.e. that last for at least 3 minutes. 

\item Extract the ANPs of each event frame and rank them by their probability ($Prob_{ANP_j}$) of describing an image. 

\item Select the top-5 ANPs per image, since we consider that those are the concepts with higher relevance, thus better capturing the image's information. After this step the model ends up with a total of \textit{M} semantic concepts per event, where $\{M =$  Number of images $\times \; 5\}$.  

\item Cluster the $M$ semantic concepts based on their Wordnet-based nouns semantic distances. As a result, we have clusters of concepts with semantic similarity. For the event sentiment computation ($S_{event}$), focus on the largest cluster. 

\item Finally, fuse the sentiment associated to the ANPs and noun's cluster following the eq. (\ref{eq:sentim}): 
\begin{equation}
\label{eq:sentim}
S_{event} =
\sum_j (\alpha *  S_{ANP_j}+\beta*S_{Noun_j}), j=1:N_{ANP}, \\ 
\vspace{-0.5em}
\end{equation}

where $S_{ANP_j} = (S_{ANP_j}^{VSO} * Prob_{ANP_j}), $ $S_{ANP_j}^{VSO}$ is the ANP's sentiment given by the VSO and $S_{Noun}$ is the label of the noun, $\alpha$ and $\beta$ are the contributions (\%) of the ANPs and the nouns. Take into account the probability associated to the ANPs aiming to penalize the ANPs with low relation to the image content. 
\end{enumerate}


\section{Experiments Setup}
\label{sec:experimentsetup}

\subsection{Dataset:}
We collected a dataset of 4495 egocentric pictures, which we call UBRUG-Senti. The user was asked to wear the Narrative Clip Camera\footnote{http://getnarrative.com/} fixed to his/her chest during several hours every day and was asked to continue with his/her normal life. Since the camera is attached to the chest, the frames vary following the user's movement and describe the user's view of his/her daily indoor/outdoor activities. It involves challenging backgrounds due to the scene variation, handled objects appearing and disappearing during images sequences, and the movement of the user. The camera takes a picture every 30 seconds, hence each day around 1500 images are collected for processing. The images have a resolution of 5MP and JPG format.

After the temporal clustering \cite{Dimiccoli2015SR-Clustering:Segmentation}, we obtained a dataset composed of 4495 images grouped in a total of 98 events. The events were manually labelled based on how the user felt while reviewing them. The labels assigned were \emph{Positive} (36), \emph{Negative} (43) and \emph{Neutral} (19). Some examples are given in Fig \ref{fig:imagesexamples}. 

\subsection{Experiments:}
During the experimental phase, we evaluated the contributions of ANPs and nouns by defining different combinations of $\alpha$ and $\beta$. We performed a balanced 5-fold cross validation. For each of the folds, we used 80\% of the total of events per label of our dataset and compute the best pair of $\alpha$ and $\beta$ values. This is a parameters selection process that is later re-evaluated in a test phase with a different set of events. 
 
\textbf{Validation:} To evaluate the effectiveness of the scene detection approach, we use the \textit{Accuracy}, as the rate of correct results, and the \textit{F-Score} (F1). The F1 is defined as : $F1= 2(RP)/(R+P)$, where $P$ is the precision $(P=TP/(TP+FP)$, $R$ is the recall $(R=TP/(TP+FN)$ and $TP$, $FP$ and $FN$ respectively are the number of true positives, false positives and false negatives of the event's sentiment label correctly identified.

\textbf{Results:} Tables \ref{trainingresults} and \ref{Testresults} present the results achieved by the proposed method at image and event level, respectively. The model achieves an average training accuracy of 73$\pm$3.8\% and F-score of 59$\pm$5.4\% and test accuracy of 75$\pm$8.2\% and F-score of 61$\pm$13.2\%, when $\alpha = 0.8$ and $\beta = 0.2$, i.e. when the ANP information is considered; although the major contribution comes from the noun sentiment associated. As expected, neutral events are the most challenging ones to classify.

\begin{table}[ht!]
\centering
\caption{Parameter-selection results}
\label{trainingresults}
\resizebox{\columnwidth}{!}{%
\begin{tabular}{cc|c|c|c|c|c}
\cline{2-7}
 & \multicolumn{3}{c|}{Accuracy} & \multicolumn{3}{c}{F-Score} \\ \cline{2-7} 
 & \begin{tabular}[c]{@{}c@{}}beta = 0.2\\ alpha = 0.8\end{tabular} & \begin{tabular}[c]{@{}c@{}}beta = 0.5\\ alpha = 0.5\end{tabular} & \begin{tabular}[c]{@{}c@{}}beta = 0.8\\ alpha = 0.2\end{tabular} & \begin{tabular}[c]{@{}c@{}}beta = 0.2\\ alpha = 0.8\end{tabular} & \begin{tabular}[c]{@{}c@{}}beta = 0.5\\ alpha = 0.5\end{tabular} & \begin{tabular}[c]{@{}c@{}}beta = 0.8\\ alpha = 0.2\end{tabular} \\ \hline
\multicolumn{1}{c|}{\textbf{Ours}} & 0.60 & 0.63 & \textbf{0.73} & 0.35 & 0.43 & \textbf{0.59} \\ \cline{1-1}
\multicolumn{1}{c|}{\begin{tabular}[c]{@{}c@{}}Evaluating\\ 3 Clusters\end{tabular}} & 0.68 & 0.66 & 0.68 & 0.48 & 0.45 & 0.48 \\ \cline{1-1}
\multicolumn{1}{c|}{\begin{tabular}[c]{@{}c@{}}Evaluating\\ with weights\end{tabular}} & 0.65 & 0.65 & 0.66 & 0.41 & 0.43 & 0.47 \\ \hline
\end{tabular}
}
\end{table}

\begin{wraptable}{r}{5.5cm}
\centering
\caption{Test set results}
\label{Testresults}
\begin{tabular}{ccc}
\cline{2-3}
 & \multicolumn{1}{c|}{Accuracy} & F-Score \\ \cline{2-3} 
 & \multicolumn{2}{c}{\begin{tabular}[c]{@{}c@{}}beta = 0.8\\ alpha = 0.2\end{tabular}} \\ \hline
\multicolumn{1}{c|}{Ours} & \multicolumn{1}{c|}{\textbf{0.75$\pm$0.08}} & \textbf{0.60$\pm$0.13} \\ \cline{1-1}
\multicolumn{1}{c|}{\begin{tabular}[c]{@{}c@{}}Evaluating\\ 3 Clusters\end{tabular}} & \multicolumn{1}{c|}{0.69$\pm$0.1} & 0.50$\pm$0.15\\ \cline{1-1}
\multicolumn{1}{c|}{\begin{tabular}[c]{@{}c@{}}Evaluating\\ with weights\end{tabular}} & \multicolumn{1}{c|}{0.74$\pm$0.1} & 0.58$\pm$0.15
\end{tabular}
\end{wraptable} 

In order to contextualize our results, we  fine-tune the well-known \textit{GoogleNet} deep convolutional neural network \cite{Ma2016GoingRecognition} to classify into Positive, Neutral and Negative. We use 80\%, 10\% and 10\% of the dataset for training, validation and testing respectively. The network achieves an accuracy of \textbf{55\%}. 

From the results we can conclude that the application of the DeepSentiBank presents drawbacks when applied to egocentric photostreams. To begin with and as commented before, the 2089 ANPs not necessarily have the power to represent what the image captured about the scene, taking into account the difficulty to detect them automatically (Mean average accuracy of the net $\sim$25\%). Moreover, the ANPs present the limitation that they are classified strictly into Negative or Positive concepts. Thus, moments from our daily routine, which are often considered as neutral, are difficult to recognize.

\section{Conclusions}
\label{sec:conclusions}

We present a new model for positive moments recognition from our digital memory, composed by images recorded by the Narrative wearable camera. 
It analyses  semantic concepts called ANPs extracted from the images. These semantic concepts have a sentiment value associated and describe the appearance of concepts in the images. The sentiment prediction tool is based on new semantic distance of ANPs and fusion of ANPs and nouns sentiments extracted from egocentric photostreams. 
The proposed approach obtained a classification accuracy of 75\% on the test set, with deviation of 8\%. 
Future experiments will address the generalization of the model over datasets collected by other wearable cameras, as well as  recorded by different users.
Analysing the results obtained, we conclude that the polarity of the ANPs makes it difficult to classify '\textit{Neutral}' events. However, most of our daily life is composed by neutral events, which can be considered as routine. Thus, in future lines we will address the routine recognition and retrieval.

\textbf{Acknowledgements:}
This work was partially founded by Ministerio de
Ciencia e Innovaci\'on of the Gobierno de Espa\~na, through the research project TIN2015-66951-C2. SGR 1219, CERCA, \textit{ICREA Academia 2014} and Grant 20141510 (Marat\'{o} TV3). The funders had no role in the study design, data collection, analysis, and preparation of the manuscript. 

\bibliographystyle{abbrv}

\end{document}